\documentclass{article}

\PassOptionsToPackage{numbers,compress}{natbib}
\usepackage{algorithm}
\usepackage{algorithmic}
 \usepackage[preprint]{neurips_2025}

\newcommand{\ReflectPretrain}{\textit{Reflection-Pretraining}}

\usepackage[utf8]{inputenc} 
\usepackage[T1]{fontenc}    
\usepackage{hyperref}       
\usepackage{url}            
\usepackage{booktabs}       
\usepackage{amsfonts}       
\usepackage{nicefrac}       
\usepackage{microtype}      
 \usepackage{xcolor}         

\usepackage[dvipsnames, svgnames, x11names]{xcolor}  
\usepackage{tcolorbox}
\usepackage{colortbl} 

\usepackage{threeparttable}
\usepackage{multirow}

\usepackage{bm}
\usepackage{CJK}
\newcommand{\ctext}[1]{\begin{CJK}{UTF8}{gbsn}\small{#1}\end{CJK}}
\usepackage{amssymb}
\usepackage{amsmath}
\usepackage{tcolorbox}
\tcbuselibrary{listings, breakable}
\usepackage{sansmath}
\usepackage{mathrsfs}
\usepackage{bbm}

\usepackage{amsmath} 
\usepackage{graphicx}
\usepackage{wrapfig}

\usepackage{mathtools}
\DeclareMathAlphabet{\mathsfit}{\encodingdefault}{\sfdefault}{m}{sl}
\SetMathAlphabet{\mathsfit}{bold}{\encodingdefault}{\sfdefault}{bx}{n}

\definecolor{lm_purple}{RGB}{227,227,240}

\title{Reflection Pretraining Enables Token-Level Self-Correction in Biological Sequence Models}

\author{
Xiang Zhang$^{1,3*}$, Jiaqi Wei$^{2,4*}$, Yuejin Yang$^{2}$, Zijie Qiu$^{1,2}$, \\ \textbf{Yuhan Chen$^{2}$,
Zhiqiang Gao$^{2}$, Muhammad Abdul-Mageed$^{3}$}, \\  \textbf{Laks V. S. Lakshmanan$^{3}$, Wanli Ouyang$^{5}$, Chenyu You$^{6}$, Siqi Sun$^{1,2}$} \\
$^1$Fudan University\qquad
$^2$Shanghai Artificial Intelligence Laboratory\qquad
$^3$University of British Columbia\qquad  \\
$^4$Zhejiang University\qquad
$^{5}$ The Chinese University of Hong Kong\qquad
$^{6}$ Stony Brook University \\
\texttt{xzhang23@ualberta.ca, siqisun@fudan.edu.cn} 
\\ \small * Equal Contribution
}

\begin{document}

\maketitle

\begin{abstract}
  \textbf{Chain of Thought (CoT)} prompting has significantly advanced task-solving capabilities in Natural Language Processing with LLMs. Unlike standard prompting, CoT encourages the model to generate \textbf{{intermediate reasoning steps}}---\textbf{non-answer tokens}---that help guide the model toward more accurate final outputs. These intermediate steps enable more complex reasoning processes such as error correction, memory management, future planning, and self-reflection.
Under appropriate assumptions, an autoregressive Transformer augmented with \textbf{natural language} (e.g., English) based  CoT can, in theory, achieve \emph{Turing completeness}, as demonstrated in prior work. However, applying CoT to \textbf{non-natural language domains}, such as protein and RNA language models, is not yet possible---primarily due to the {limited expressiveness} of their token spaces (e.g., amino acid tokens).
 In this work, we propose and define the concept of \emph{language expressiveness},  the ability of a given language using its tokens as well as its grammar to encode various information. We show that the limited expressiveness of protein language severely restricts the applicability of CoT-style reasoning. To overcome this, we introduce \textbf{reflection pretraining}---for the first time in a biological sequence model---which enables the biological model to engage in intermediate reasoning through the generation of auxiliary ``thinking tokens'' beyond simple answer tokens.
Theoretically, we demonstrate that our augmented token set significantly enhances the \textbf{biological language expressiveness}, thereby improving the overall reasoning capacity of the model. Experimentally, our novel pretraining approach teaches protein models to \emph{self-correct} and leads to substantial performance gains compared to standard pre-training. 
Finally, we show that reflection training brings unique advantages, such as improved resistance to overfitting (i.e., \emph{counter-memorization}) and enhanced \textbf{human steer-ability}---enabling users to interfere/interact with the protein generation---thus bridging the gap between {biological language models} and human {natural language models}. \textbf{All code, trained model weights, and result outputs} are publicly available on our \href{https://github.com/BEAM-Labs/denovo/}{\color{blue}\textit{GitHub repository}}. Detailed theoretical analysis, discussions on model expressiveness,  extensive experimental results, and related work section are provided in the Appendix.

\end{abstract}

\section{Introduction}
Deep learning~\cite{lecun2015deep} has significantly advanced the field of biology, with an increasing number of neural models being trained to generate and predict biological sequences such as DNA\cite{alipanahi2015predicting,zhang2021deep,rizzo2015deep,lo2016deep}, RNA~\cite{yang2022scbert,deng2022rapid,tian2021model}, and proteins~\cite{xiao2025protein,rives2021biological,rao2021msa,zhang2025bidirectional,jumper2021highly,lin2023evolutionary,guntuboina2023peptidebert,elnaggar2021prottrans,meier2021language,brandes2022proteinbert,wu2021protein,brandes2022proteinbert,yang2024introducing,xia2024adanovo,eloff2023novo,yilmaz2022novo,zhang2025language,zhangdistilling,qiuranknovo}. However, current biological sequence-generation models are constrained to produce only \textbf{answer tokens} directly related to specific tasks (e.g., drug design, de novo sequencing~\cite{jin2024contranovo,zhang2025pi,qiu2025universal}). This generation paradigm mirrors conventional natural language processing models~\cite{naveed2023comprehensive}, where outputs are limited to final answers without intermediate reasoning or deliberation.\footnote{For example, a machine translation model  outputs just``goodbye'' for the input ``au revoir,'' or a mathematical language model  returns just answer ``121'' for the input ``11 $\times$ 11,'' without revealing intermediate steps.} Recent work~\cite{wei2022chain,zhang2024autoregressive+,zhang2024counting} has demonstrated that this answer-only generation approach is suboptimal, both in terms of theoretical expressiveness and empirical performance. While a full theoretical analysis is provided in the {\color{blue} Appendix}, the core intuition is straightforward: solving complex tasks, especially those requiring reasoning, often involves iterative exploration, including trial-and-error, partial hypotheses, and even initial incorrect outputs before arriving at a final solution. Models constrained to generate only final answers are fundamentally incapable of performing this kind of structured, exploratory computation and thus fail to handle complex solution discovery effectively.

Chain-of-Thought (CoT)~\cite{wei2022chain,hu2025survey} prompting fundamentally changes how answers are generated in natural language models. Traditional neural models directly map an input sequence to a sequence of \textbf{answer tokens}, expressed as:
$
\bm{x_i:x_n} \Rightarrow \texttt{<answer}_1\texttt{>} \ \texttt{<answer}_2\texttt{>} \cdots \texttt{<answer}_m\texttt{>}
$.

In contrast, CoT introduces interleaved \textbf{non-answer tokens} that enable intermediate reasoning~\cite{wei2022chain}: 
\[ 
x_i:x_n \Rightarrow \texttt{<answer}_1\texttt{>} \ [\texttt{non-answer}_1] [\texttt{non-answer}_2] \ \texttt{<answer}_2\texttt{>} \cdots [\texttt{non-answer}_k] \ \texttt{<answer}_m\texttt{>}.
\] 
Although these non-answer tokens are discarded in the final output, they significantly enhance the model's capabilities by enabling it to \textit{store intermediate memory, perform iterative computation, correct earlier errors, and reason across multiple steps}.

This augmentation enables natural language (e.g. English) models to perform human-like reasoning. Under suitable assumptions, CoT-enhanced models can even approach the theoretical upper bound of Turing completeness~\cite{li2024chain} (detailed in {\color{blue} Appendix}), a level of expressiveness unattainable by direct-answer generation alone~\cite{deletang2022neural}.

Theoretically, the increase in computational power is partly attributed to CoT's capacity for \textit{vector-to-token conversion}, which enables the simulation of recurrent computation~\cite{zhang2024autoregressive+} (see {\color{blue} Appendix}). Moreover, CoT allows for unbounded generation of \textbf{non-answer tokens}, effectively serving as an "unlimited" memory tape in the ideal case~\cite{li2024chain,wei2025alignrag}. Crucially, such a process is only feasible due to the expressive capacity of human natural languages (e.g., English), which we formalize under the concept of \emph{language expressiveness}.

Highly expressive languages can encode rich, structured information from hidden states $\mathbf{h}$ into tokens, including reflections on prior errors, intermediate computations (e.g., counters, partial products), perceptual attributes (e.g., RGB colors), and procedural knowledge like algorithms. 
In contrast, language like protein~\cite{rives2021biological,yilmaz2022novo,wei2025ai} or RNA~\cite{yang2022scbert} relies on token and grammar systems with limited expressiveness, restricting models to output only the next amino acid or nucleotide rather than diverse information used in the CoT process. \textbf{In this paper, we formally demonstrate that the expressiveness of the foundational language directly affects the theoretical upper bound on the model’s overall reasoning capability.} Limited bio. language's expressiveness resulted in the gap between Natural Language Models (e.g., GPT~\cite{achiam2023gpt}) and Bio Language models (ESM~\cite{rives2021biological}, Prollama~\cite{lv2025prollama}).

In this work, we augment biological sequence models with \textit{non-answer reasoning tokens} through a novel reflection-based pretraining method. Focusing on the task of \textit{de novo} peptide (short protein) sequencing, we conduct large-scale pretraining on a carefully curated reflection dataset, introducing two mechanisms for injecting prediction errors and teaching the model to self-reflect and self-correct.

\textbf{Our extensive experiments yield the following key findings}:  
\textbf{(1)} With appropriate training techniques and data augmentation, protein sequence models can perform \textbf{intermediate reasoning, self-correction, and reflection} capabilities that go beyond simply generating just answer tokens;   
\textbf{(2)} Finetuning alone does not confer reasoning ability in bio-models; such capacity \textbf{emerges only through  pretraining};  
\textbf{(3)} Within a certain threshold, increasing the number of manually injected errors during training improves the model’s ability to  \textbf{self-correct};   
\textbf{(4)} Reflection pretraining leads to significant accuracy gains, driven by both \textbf{enhanced thinking ability} and a  \textbf{counter-overfitting} effect during training;   
\textbf{(5)} The resulting models support \textbf{human-in-the-loop generation}, enabling biologists to interact with the CoT process. This narrows the gap between biological language models and large (natural) language models.

\section{Bio Language Expresiveness}
\label{sec:Bio Language Expresiveness}

\subsection{Model computing power and relations to languages}
Theoretical investigations~\cite{zhang2024autoregressive+,deletang2022neural,li2024chain} into the computational power of neural models--that is, the class of problems a given architecture can solve--have been extensively explored. Models constrained to output only \textbf{answer tokens}, without any intermediate outputs, are known to possess limited expressiveness, often corresponding to low-complexity classes such as \textsf{TC}$_0$ or \textsf{TC}$_1$~\cite{li2024chain,zhang2024autoregressive+,deletang2022neural}.

\begin{figure}[h]
    \centering
    \includegraphics[width=0.75\textwidth]{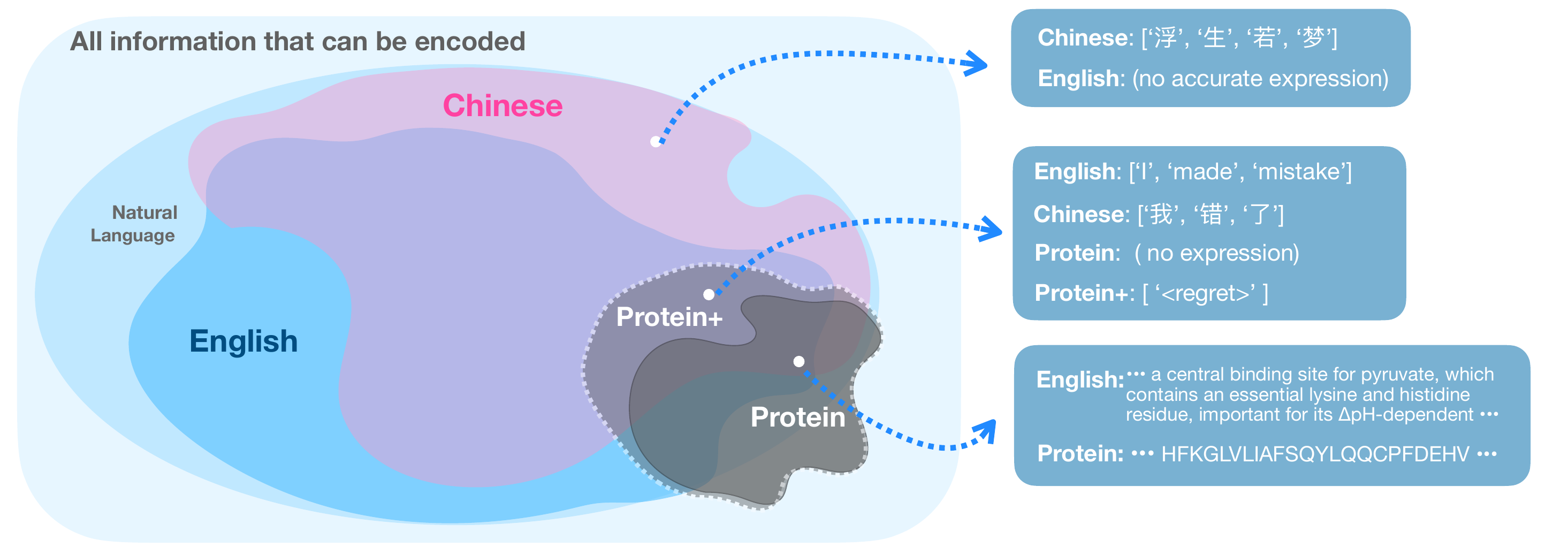} 
    \caption{Expressiveness of Languages. }
    \label{fig:expressinvess}
\end{figure}

While the full formal analysis is provided in the {\color{blue} Appendix}, the intuition is straightforward. First, such models suffer from shallow reasoning depth due to their non-recurrent architectures and the limited number of output steps--answers are short and final by design. Second, their overall computational complexity is constrained, as the time available for reasoning is tightly bound by the short answer sequence. Third, the lack of intermediate outputs limits memory capacity, restricting the model’s ability to store and refer to intermediate results during computation. These combined constraints fundamentally limit the model's ability to simulate complex, multi-step algorithms.

Chain-of-Thought (CoT) mitigates all three aforementioned limitations by introducing auxiliary \textit{reasoning tokens}--tokens that are not part of the final answer but assist in reaching it. Theoretical and empirical studies on CoT have demonstrated that this process increases computational \textbf{depth} through vector-to-token conversions, allowing intermediate information to propagate over time steps; enhances computational \textbf{complexity} by allocating additional computation before emitting answer tokens; and expands effective \textbf{memory} by storing information in intermediate natural language tokens that act as an unbounded memory tape.

However, existing analyses of CoT’s computational power are exclusively grounded in \textbf{natural language} models, which is the dominant focus of the deep learning community. These studies implicitly rely on a critical but often overlooked assumption: the expressive capacity of natural language tokens--their ability to encode and convey diverse types of information. In this section, \textbf{we formally introduce the concept of language expressiveness} and analyze its fundamental role from different languages (English v.s. Protein) in determining the theoretical limits of model design.
\subsection{Language Expressiveness}
We define the expressiveness of any given language as follows: 
\begin{tcolorbox}[colback=gray!10, colframe=gray!60, title=Definition: Language Expressiveness]
A language is defined as $\bm{L} = (\mathcal{G}, \mathcal{V})$, where $\mathcal{G}$ denotes the language's grammar specifying valid token sequences, and $\mathcal{V} = \{ \texttt{t}_1, \texttt{t}_2, \dots, \texttt{t}_M \}$ is the vocabulary of usable tokens in this language. The set of all meanings expressible (valid sequences) in language $\bm{L}$ is denoted $\mathbb{S}_{\bm{L}}$. \\

For any syntactically valid token sequence $\bm{s} = (\texttt{t}_{i_1}, \texttt{t}_{i_2}, \dots)$, grounded in $\mathcal{G}$, it encodes certain information (meaning) $\bm{s} \in \mathbb{S}_{\bm{L}}$. The \textbf{expressiveness} of $\bm{L}$ is defined as the cardinality:
\[
\textnormal{Expressiveness}(\bm{L}) := \left|\mathbb{S}_{\bm{L}}\right|.
\]
\end{tcolorbox}

While amount of information a language can encode, i.e. expressiveness $\left|\mathbb{S}_{\bm{L}}\right|$, is often infinite (i.e.$|\mathbb{S}_{\bm{L}}| = \infty$ as there's countable infinite sequences in a language and each sequence defines some information $s$), languages can still be meaningfully compared. If $\mathbb{S}_{\bm{L}_1} \subset \mathbb{S}_{\bm{L}_2}$ and $\mathbb{S}_{\bm{L}_1} \neq \mathbb{S}_{\bm{L}_2}$, we say that $\bm{L}_2$ is strictly \textbf{more expressive} than $\mathbb{L}_1$. For instance, $|\mathbb{S}_{\textnormal{English}}| \subset $ $|\mathbb{S}_{\textnormal{NaturalLanguage}}|$ since English is one of natural language (Fig.~\ref{fig:expressinvess}). 

Natural languages, such as English and Chinese, possess high expressiveness due to large vocabularies and flexible compositional grammar structures. They are often not directly \textit{comparable} in expressiveness, as each may contain idioms or cultural references untranslatable by the other. This yields $\mathbb{S}_{\textnormal{English}} \not\subseteq \mathbb{S}_{\textnormal{Chinese}}$ and $\mathbb{S}_{\textnormal{Chinese}} \not\subseteq \mathbb{S}_{\textnormal{English}}$, but $\mathbb{S}_{\textnormal{English}} \cap \mathbb{S}_{\textnormal{Chinese}} \neq \emptyset$ (Fig.~\ref{fig:expressinvess}).

\begin{figure}[!h]
    \centering
    \includegraphics[width=\textwidth]{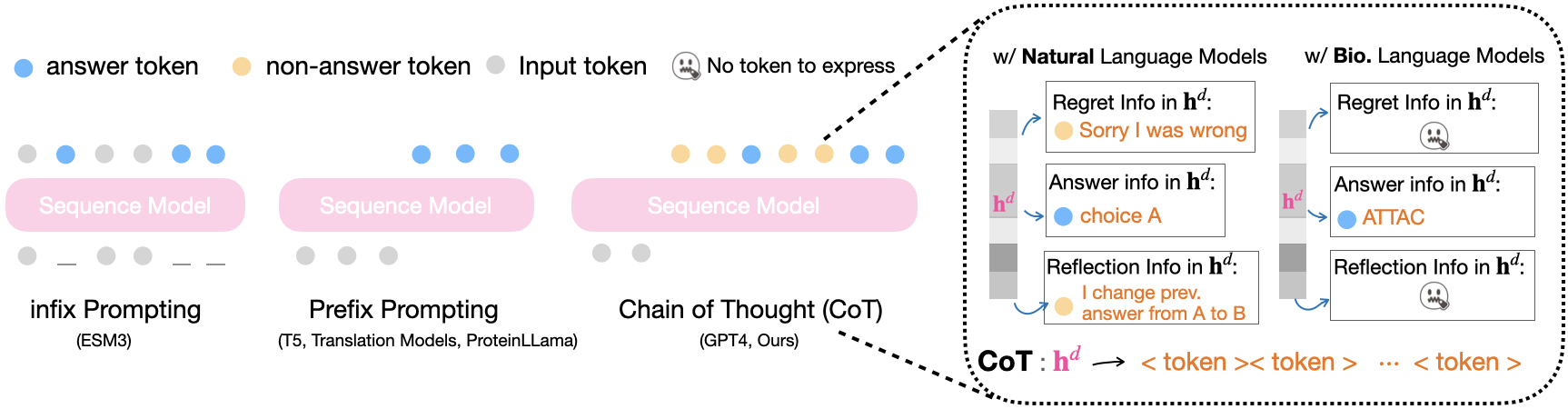} 
    \caption{Comparative framework for prompting in natural and biological language models. }
    \label{Fig:cot+language}
\end{figure}

Biological sequence models--such as protein models--can also be formalized as languages: $\bm{L}_{\textnormal{protein}} = (\mathcal{G}_{\textnormal{protein}}, \mathcal{V}_{\textnormal{protein}})$, where $\mathcal{V}_{\textnormal{protein}}$ comprises the 20 amino acids. Despite functioning as a language for encoding biological information, its expressiveness is severely limited. For example, sequences representing \texttt{regretful} information such as $\bm{s}_{\textnormal{en}} = [$\texttt{``I''}, \texttt{``am''}, \texttt{``wrong''}$]$ in English and $\mathbf{s}_{\text{zh}} = [\text{``\ctext{我}''}, \text{``\ctext{错}''}, \text{``\ctext{了}''}]$ in Chinese,  have no meaningful expression in $\mathbb{S}_{\textnormal{protein}}$ (Fig.~\ref{fig:expressinvess}). This limitation obstructs the representation of reasoning, posing a significant barrier to applying CoT-style methods in protein-based models (Fig.~\ref{Fig:cot+language}  Right ).

\subsection{Language expressiveness determines CoT expressiveness}
Chain-of-Thought (CoT) reasoning enables models to generate not only answer tokens, denoted as $\langle \texttt{a} \rangle \in \mathcal{V}_{\bm{L}}$, but also non-answer tokens $\langle \texttt{na} \rangle$ drawn from the same vocabulary. In contrast, standard prompting--weather via infix or prefix format (Fig.~\ref{Fig:cot+language}), as seen in traditional NLP systems and all current biological sequence models--maps input prompt tokens directly to output answers, $\bm{x_1}:\bm{x_n} \Rightarrow \langle \texttt{a}_1 \rangle \langle \texttt{a}_2 \rangle \cdots \langle \texttt{a}_k \rangle$ (Fig. \ref{Fig:cot+language}).This direct generation scheme imposes computational limitations, as the number of answer tokens is inherently constrained, thereby capping the total computation that the model can express (see {\color{blue} Appendix}). In contrast, CoT introduces a richer computational structure by interleaving non-answer tokens throughout the generation process:
$\bm{x}_1 : \bm{x}_n \Rightarrow [\texttt{na}_1] [\texttt{na}_2] \langle \texttt{a}_1 \rangle [\texttt{na}_3] [\texttt{na}_4] \langle \texttt{a}_2 \rangle \cdots \langle \texttt{a}_k \rangle
$ 
(Fig.~\ref{Fig:cot+language}). These intermediate tokens act as memory buffers and reasoning steps, allowing models to conduct deeper computation before committing final answers.

The $\langle \texttt{na} \rangle$ tokens generated during Chain-of-Thought (CoT) reasoning are \textbf{not randomly generated}. Rather, they are discretized, verbalized, tokenized, or quantized forms of internal information encoded in the model's hidden state vector $\mathbf{h}$. neural models store \textbf{all} computational information-such as counters, memory, variables, and intermediate results-in this latent vector, typically of fixed dimensionality (e.g.512).

If we define the latent space as language $\bm{\Theta} = (\mathcal{G}_\theta, \mathcal{V}_\theta)$, where $\theta$ is learned parameter from model. Essentially, information latent language encoded, $\mathbb{S}_\theta$, is a set of information that all latent vectors $\mathbf{h}$ can encode after model $\theta$ is trained. Naturally, a better-trained model expresses richer information and expressiveness $|\mathbb{S}_{\bm{\Theta}}|$ will be bigger.

\tcbset{
  colback=gray!10, 
  colframe=gray!80, 
  boxrule=0.5pt,
  arc=2pt,
  left=6pt,
  right=6pt,
  top=4pt,
  bottom=4pt
}

\begin{tcolorbox}[title=\textbf{Essence of CoT}]

Each step of CoT can be viewed as a mapping from the latent language space to natural language: $\mathbb{S}_{\bm{\Theta}} \rightarrow \mathbb{S}_{\bm{L}}$, where the latent state $\mathbf{h} \in \mathcal{V}_\theta$ is decoded into a sequence of non-answer language tokens $([\texttt{na}_1], [\texttt{na}_2], \dots, [\texttt{na}_k])$, with $[\texttt{na}_k] \in \mathcal{V}_{\bm{L}}$ (Fig.~\ref{Fig:cot+language}, right panel).

\end{tcolorbox}
For example, the model's  ``\textit{regret}'' information encoded in $\mathbf{h}$ can be expressed as the token sequence [\texttt{'I'}, \texttt{'made'}, \texttt{'a'}, \texttt{'mistake'}] during English-based CoT ((Fig.~\ref{Fig:cot+language}). Similarly, a latent representation corresponding to ``\textit{not sure of the answer yet}'' may be verbalized as [\texttt{'ummm'}] to facilitate further computation. Conversely, ``\textit{confident}'' information in latent states are often decoded directly into final answer tokens.

Due to the \textbf{limited expressiveness} of bio. language such as protein $\bm{L}_\textnormal{protein}$, the latent information in vectors $\mathbf{h}$ cannot be explicitly expressed in language space using amino acids ($\mathcal{V}_\textnormal{protein}$) and its grammar ($\mathcal{G}_\textnormal{protein}$) as shown in Fig.~\ref{Fig:cot+language}, right panel, unlike in natural language-based LLMs. \textbf{As a result, the CoT expressiveness with Biological sequence models is practically $0$}.

\begin{figure}[!h]
    \centering
    \includegraphics[width=\textwidth]{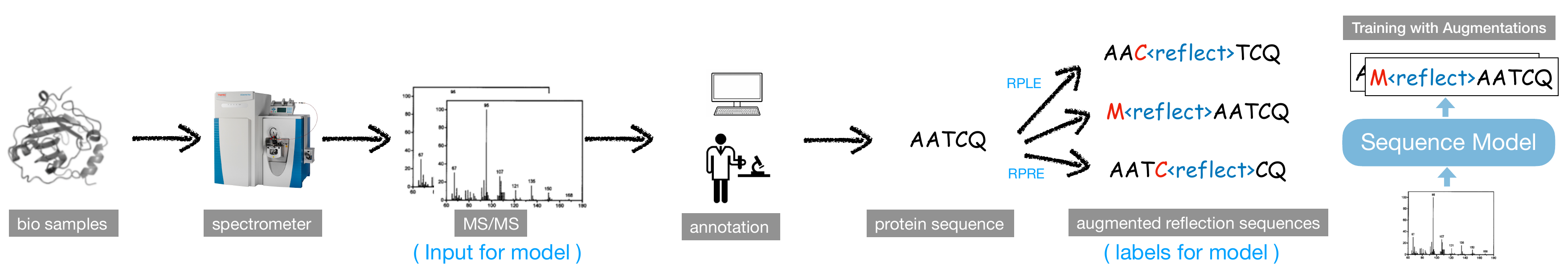} 
    \caption{De novo peptide sequencing workflow using tandem mass spectrometry.}
    \label{Fig:denovo}
\end{figure}

\section{Reflection pretraining for protein (peptide) sequence predictions}

\subsection{Task Choice and Problem Formulation}
To this end, we propose \textit{reflection-pretraining} for protein sequencing models to allow CoT reasoning beyond generating solely answer tokens (amino acids). We choose the the task of \textbf{de novo peptide sequencing} (Fig. \ref{Fig:denovo}) for five key reasons:  
1) It is a foundational problem in protein sequence prediction, as peptide sequencing remains the primary method for determining amino acid sequences from nature.  
2) The task is reasoning-intensive, requiring the model to perform complex computations on spectrum signals-well aligned with the reflection paradigm.  
3) Large-scale training data is readily available.  
4) Evaluation is precise, with each input spectrum paired with a gold-standard sequence, unlike protein language modeling with ambiguous metrics.  
5) Each token corresponds to a discrete reasoning step, enabling fine-grained, token-level reflection.

In the \textit{de novo} peptide sequencing task (Fig. \ref{Fig:denovo}), the model is given a spectrum instance $\mathbf{H} = \{\mathbf{I}, c, m\}$, produced by a mass spectrometer (Fig. \ref{Fig:denovo} when provided with a biological sample (Fig. \ref{Fig:denovo}. The spectrum consists of:

\begin{itemize}
    \item $\mathbf{I} = \{(\text{m/z}_1, i_1), (\text{m/z}_2, \text{i}_2), \dots, (\text{m/z}_k, \text{i}_k)\}$, a set of $k$ observed mass-to-charge and intensity pairs, filtered by a signal threshold, (Fig. \ref{Fig:denovo} MS/MS)
    \item $c \in \mathbb{Z}^+$, the charge state of the precursor ion, and
    \item $m \in \mathbb{R}^+$, the total measured mass of the peptide.
\end{itemize}

The objective is to \textbf{predict the underlying amino acid (protein) sequence} $\mathbf{A} = \{a_1, a_2, \dots, a_n\}$, where each token $a_i \in \mathcal{V}_\textnormal{protein} = \{ \textnormal{20 amino acids tokens}\}$ belongs to the vocabulary of the protein language $\bm{L}_\textnormal{protein} = (\mathcal{G}_\textnormal{protein}, \mathcal{V}_\textnormal{protein})$. The mapping $\mathbf{H} \mapsto \mathbf{A}$ requires the model to reason over the structure and intensity patterns in $\mathbf{I}$ while conforming to biochemical constraints imposed by $c$ and $m$.

\textit{Note that, although we describe our method in the context of the protein sequence space, it applies to other biological sequence prediction tasks-including RNA, DNA, and synthetic polymers}.

\subsection{Reflection Token augmented Protein sequence for next token prediction training}
One of the most powerful forms of reasoning is self-reflection and error correction. Prior work shows that sequence models often encode a sense of ``regret'' in their latent state $\mathbf{h}$-that is, the realization of an error \textit{after} generating an answer token $\langle \text{a} \rangle$. In large language models (LLMs), the high expressiveness of the English language $\bm{L}_\text{english}$ enables this information to be verbalized, making reflection a highly effective mechanism for improving reasoning accuracy. However, in protein models, this is not possible: the latent regret state $\mathbf{h}_\text{regret}$ cannot be decoded into any token $a \in \mathcal{V}_\text{protein}$, i.e., $\mathbf{h}_\text{regret} \Rightarrow a$ is infeasible.

To enable self-reflection in protein sequence models, we augment the vocabulary with a reflection token, defining $\mathcal{V}_\text{protein+} = \mathcal{V}_\text{protein} \oplus \langle \texttt{reflect} \rangle$. We then construct a modified training dataset $\mathcal{D}_\text{de novo+}$ (Figure \ref{Fig:denovo} ) by injecting synthetic errors (e.g., incorrect amino acids) into sequences and appending corrections using $\langle \texttt{reflect} \rangle$.

\begin{figure}[!h]
    \centering
    \includegraphics[width=0.9\textwidth]{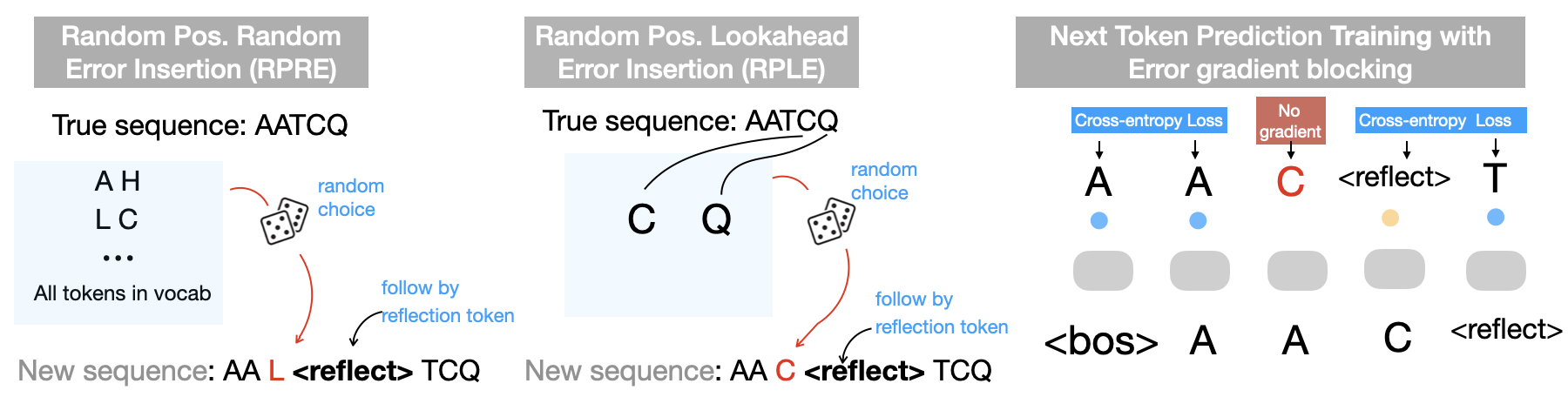} 
    \caption{Error injection and reflection training for augmenting reasoning in bio. sequence models.}
    \label{Fig:error}
    \vspace{-0.8em}
\end{figure}

In this section, we introduce two novel strategies for error injection (Fig. \ref{Fig:error}), as well as our dynamic data updating and gradient blocking strategy for training protein models with reflection-based next-token prediction on the \textit{de novo} sequencing task. 

\subsubsection{Model Architecture}

We adopt a Transformer-based encoder-decoder architecture, following standard designs for this task~\citep{yilmaz2023sequence}. The input spectrum is denoted as $\bm{H} = \{\mathbf{I}, c, m\}$, where $\mathbf{I} = \{(m/z_k, i_k)\}_{k=1}^K$ consists of $K$ mass-to-charge and intensity pairs. Each $(m/z_k, i_k)$ is embedded via a sinusoidal encoding and projected into a $d$-dimensional vector. The resulting sequence of embeddings forms the encoder output $\bm{E} \in \mathbb{R}^{K \times d}$.

The decoder is an autoregressive Transformer with causal self-attention and cross-attention to $\bm{E}$. Let $\bm{h}_t^{(i)}$ denote the hidden state at decoding step $t$ and decoder layer $i$. Causal self-attention ensures that each position only attends to past outputs:
$\bm{h}_t^{(i)} = \textnormal{Attn}\left(\bm{h}_t^{(i-1)}, \{\bm{h}_{1:t-1}^{(i-1)}\}\right)
$.
Each $\bm{h}_t^{(i)}$ then attends to the spectrum features via cross-attention over $\bm{E}$. The final decoder output at step $t$, denoted $\bm{h}_t^{(L)}$, is projected onto the vocabulary $\mathcal{V}_\textnormal{protein+}$:
$
P_t(\cdot \mid \bm{S}, \bm{y}_{<t}) = \textnormal{softmax}\left(\bm{W} \bm{h}_t^{(L)}\right)
$; 
Token $\bm{y}_t$ is sampled from $P_t(\cdot \mid \bm{S}, \bm{y}_{<t})$. Training proceeds via standard next-token prediction (NTP), minimizing the cross-entropy loss over ground-truth sequences.

 \subsubsection{Random Position, Random Error (RPRE) Injection}

For a given input-spectrum and ground-truth sequence pair $(\bm{H}, \bm{A} = \{ a_1, a_2, \dots, a_n \}) \in \mathcal{D}_\textnormal{de novo}$, we perform \textit{reflection error injection} (RPRE) to construct training sequences for teaching the model how to reflect on its past generation and correct potential mistakes.

In the RPRE strategy, we randomly sample a position $t \sim \textnormal{Uniform}(1, n)$ and a random token $\tilde{a}_t \sim \mathcal{V}_\textnormal{protein}$. We then replace the token at position $t$ with $\tilde{a}_t$, regardless of correctness, and insert a reflection token $\langle \texttt{reflect} \rangle$ immediately afterward, followed by the original correct token $a_t$. The modified training target becomes:
\[
\bm{A}' = \{ a_1, \dots, a_{t-1}, \tilde{a}_t, \langle \texttt{reflect} \rangle, a_t, a_{t+1}, \dots, a_n \}
\]

This approach enables the model to learn how to self-correct through reflection. Notably, when $\tilde{a}_t = a_t$ (i.e., the random token happens to be correct), the reflection becomes a no-op-teaching the model to maintain confidence when no error is present. As a result, RPRE encourages both correction and retention behavior through the same mechanism.

\subsubsection{Random Position, Lookahead Error (RPLE) Insertion}

While RPRE introduces reflection opportunities, the injected error $\tilde{a}_t$ may be overly implausible (e.g., ``$1 + 1 = 3 \langle  \text{reflect} \rangle 2$'') and thus trivially detectable. To inject more cognitively challenging errors, we propose \textit{Random Position, Lookahead Error (RPLE)} insertion, which samples replacement tokens within the ground-truth sequence itself.

Given a spectrum-label pair $(\bm{H}, \bm{A} = \{ a_1, a_2, \dots, a_n \}) \in \mathcal{D}_\textnormal{de novo}$, we randomly select a position $t \sim \textnormal{Uniform}(1, n)$ and choose an error token $\tilde{a}_t$ from a set of true tokens that appear later in the sequence:
$
\tilde{a}_t \sim \{ a_{t+1}, a_{t+2}, \dots, a_n \}
$. 
We then replace $a_t$ with $\tilde{a}_t$, append the reflection token $\langle \texttt{reflect} \rangle$, and restore the correct token $a_t$. This strategy creates more realistic, sequence-consistent errors and helps the model learn to correct subtle confusions such as token swaps or premature token use--common in peptide prediction (e.g., predicting ``BA'' instead of ``AB'').

\subsubsection{batch-level Online Dynamic Error Injection}

To prevent the model from memorizing the same sequence and error type and encourage generalizable reflection behavior, we adopt \textit{batch-level online dynamic error injection} during training. For each mini-batch randomly sampled from $\mathcal{D}_\textnormal{de novo}$, we apply error injection in real-time to the target sequences.

Given an injection ratio $\alpha \in [0, 1]$, a fraction $\alpha$ of sequences in each batch are modified using either the RPRE or RPLE strategy. The remaining $(1 - \alpha)$ fraction are left unchanged: 

\begin{algorithm}[H]
\caption{Online Dynamic Reflection-Error Injection During Training}
\label{alg:dynamic-error}
\begin{algorithmic}[1]
\REQUIRE Batch $\mathcal{B} = \{(\bm{H}^{(i)}, \bm{A}^{(i)})\}_{i=1}^B$, injection ratio $\alpha \in [0, 1]$
\FOR{each $(\bm{H}, \bm{A})$ in $\mathcal{B}$}
    \IF{Uniform$(0,1) < \alpha$}
        \STATE Sample $t \sim \textnormal{Uniform}(1, |\bm{A}|)$, choose RPRE or RPLE
        \STATE $\tilde{a}_t \sim$
        \textbf{RPRE}: $\mathcal{V}_\textnormal{protein}$ \quad \textbf{RPLE}: $\{a_{t+1}, \dots, a_n\}$
        \STATE $\bm{A}' \gets \{ a_1, \dots, a_{t-1}, \tilde{a}_t, \langle \texttt{reflect} \rangle, a_t, \dots, a_n \}$
        \STATE Replace $\bm{A}$ with $\bm{A}'$ in $\mathcal{B}$
    \ENDIF \ENDFOR

\end{algorithmic}
\end{algorithm}

\subsubsection{Error Position Gradient Blocking}

To ensure that injected errors act as \textit{contextual signals} rather than learning targets, we apply \textbf{gradient blocking} at error positions during training.
Formally, given a modified target sequence $\bm{A}' = \{ a_1, \dots, \tilde{a}_t, \langle \texttt{reflect} \rangle, a_t, \dots \}$, we exclude the loss term at position $t$--corresponding to the injected error token $\tilde{a}_t$--from the training objective. This allows the model to condition on the error via causal attention and learn to generate the reflection token $\langle \texttt{reflect} \rangle$ in response, without learning the generation of the incorrect prediction itself (Fig. \ref{Fig:error} right panel).

\section{Experiments}

\subsection{Experimental Setup}

\noindent \textbf{Datasets.}  
Following established protocols~\citep{yilmaz2023sequence}, we use the MassIVE-KB dataset~\citep{wang2018assembling}, which contains \textbf{30 million peptide-spectrum matches (PSMs)} collected from diverse mass spectrometry platforms. Evaluation is conducted on the widely adopted 9-species-v1 and 9-species-v2 benchmarks. 

\textbf{Implementation.}  
Input peaks and amino acids are embedded into 512-dimensional vectors. Both the spectrum encoder and peptide decoder are 9-layer Transformers with 8 attention heads and 1024-dimensional hidden states. Models are trained on eight NVIDIA A100 80GB GPUs using AdamW ($\alpha_0 = 5 \times 10^{-4}$) with a 10k-step linear warmup followed by cosine decay.

\textbf{Evaluation Metrics.}  
We report two metrics. \textbf{Amino acid-level (AA) Precision} considers a prediction correct if the amino acid token aligns with the target position. Accuracy is defined as $\bm{M}_{\texttt{AA}} / \bm{T}_{\texttt{AA}}$, where $\bm{M}_{\texttt{AA}}$ is the number of matched amino acids and $\bm{T}_{\texttt{AA}}$ is the total number of predicted amino acids. \textbf{Peptide-level precision} considers a prediction correct only if the entire predicted peptide matches the ground truth sequence exactly. Precision is defined as $\bm{M}_{\texttt{pep}} / \bm{T}_{\texttt{pep}}$, where $\bm{M}_{\texttt{pep}}$ is the number of matched peptides and $\bm{T}_{\texttt{pep}}$ is the total number of predicted peptides.

\subsection{Results}
\paragraph{Results Analysis.}
As shown in Table~\ref{tab:1}, reflection-based pretraining leads to substantial improvements in both amino acid and peptide precision across all 9 species. In contrast, \textbf{finetuning } shows negligible gains over baselines, and reflection tokens are never used during inference (Table~\ref{tab:reflection-usage}), indicating that reflection behavior is not learned through finetuning alone.

Notably, increasing the error ratio from 60\% to 90\% further boosts peptide precision in all species, while also increasing the frequency of reflection token usage from 2.3\% to 4.46\% during inference (Table~\ref{tab:reflection-usage}). Despite most \texttt{<reflect>} tokens retaining the original answer (e.g., 67.8\% in the 90\% error model), this indicates a learned capacity for self-assessment.
Moreover, incorporating RPLE during pretraining introduces more realistic, confusing errors, pushing peptide precision further up.

\begin{table*}[!t]
  \centering
  \caption{
  Comparison of different settings of reflection pretraining and baseline on 9-Species test set. 
  }
  \setlength{\tabcolsep}{0.6mm}
  \renewcommand{\arraystretch}{1.1}
  \resizebox{1.0\textwidth}{!}{
  \begin{tabular}{l|ccccccccc}
    \toprule
    \textbf{Method (beam size = 1)} & \textit{Mouse} & \textit{Human} & \textit{Yeast} & \textit{M.mazei} & \textit{Honeybee} & \textit{Tomato} & \textit{R.bean} & \textit{Bacillus} & \textit{C.bacteria} \\
    \midrule
    \textbf{Standard Pretrain} & \multicolumn{8}{c}{\textbf{Amino \  Acid  \ Precision}} \\
    \quad Transformer {\color{ForestGreen}($\Delta$ baseline) }   & 0.717& 0.649 & 0.752 & 0.713 & 0.706 & 0.763 & 0.714 & 0.753 & 0.66 \\
    \textbf{Reflect. {\color{orange}Finetune} w/ \textit{RPRE}} \\
    \quad Transformer + 60\% Error & 0.737 & 0.672 & 0.736 & 0.739 & 0.690 & 0.784 & 0.749 & 0.774 & 0.684 \\
    \textbf{Reflect. {\color{cyan}Pretrain} w/ \textit{RPRE}} \\
    \quad Transformer + 60\% Error   
    & 0.765 & 0.713 & 0.796 & 0.772 & 0.721 & 0.806 & 0.795 & 0.811 & 0.713 \\
    \textbf{Reflect. {\color{cyan}Pretrain} w/ \textit{RP(RE +LE)}} \\
    \quad Transformer + 60\% Error 
      & 0.784  & 0.735  & 0.808  & 0.786  & 0.743  & 0.813  & 0.808  & 0.820  & 0.723  \\
    \quad \quad\quad\quad\quad\ \ \ \ \  + 90\% Error 
      & \cellcolor{gray!10}\textbf{0.792} & \cellcolor{gray!10}\textbf{0.752}  & \cellcolor{gray!10}\textbf{0.809 } & \cellcolor{gray!10}\textbf{ 0.790}  & \cellcolor{gray!10}\textbf{0.744}  & \cellcolor{gray!10}\textbf{0.822}  & \cellcolor{gray!10}\textbf{0.817}  & \cellcolor{gray!10}\textbf{0.826} & \cellcolor{gray!10}\textbf{0.737} \\
      \quad 
      &  \cellcolor{gray!10}({\color{ForestGreen}{+10.5\%}}) &  \cellcolor{gray!10}({\color{ForestGreen}{+15.9\%}}) & \cellcolor{gray!10}({\color{ForestGreen}{+7.6\%}}) & \cellcolor{gray!10}({\color{ForestGreen}{+10.8\%}}) & \cellcolor{gray!10}({\color{ForestGreen}{+5.4\%}}) & \cellcolor{gray!10}({\color{ForestGreen}{+7.7\%}}) & \cellcolor{gray!10}({\color{ForestGreen}{+14.4\%}}) &  \cellcolor{gray!10}({\color{ForestGreen}{+9.7\%}}) & \cellcolor{gray!10}({\color{ForestGreen}{+11.2\%}}) \\
    \quad \quad\quad\quad\quad\ \ \ \ \  + 99\% Error 
      & 0.786  & 0.739  & 0.807 & 0.789  & 0.742  & 0.819  & 0.814  & 0.823 & 0.732  \\
    \midrule
    \textbf{Standard Pretrain} & \multicolumn{8}{c}{\textbf{Peptide \   Precision}}   \\
    \quad Transformer {\color{ForestGreen}($\Delta$ baseline) }  & 0.443 & 0.433 & 0.584 & 0.522 & 0.460 & 0.606 & 0.652 & 0.580 & 0.413 \\
    \textbf{Reflect. {\color{orange}Finetune} w/ \textit{RPRE}} \\
    \quad Transformer + 60\% Error   & 0.443 & 0.434 & 0.589 & 0.530 & 0.465 & 0.607 & 0.563 & 0.583 & 0.416 \\
     \textbf{Reflect. {\color{cyan}Pretrain} w/ \textit{RPRE}} \\
    \quad Ours (60\% ER, 6 AA/PEP)   
    & 0.485 & 0.494 & 0.632 & 0.575 & 0.507 & 0.644 & 0.629 & 0.635 & 0.458 \\
     \textbf{Reflect. {\color{cyan}Pretrain} w/ \textit{RP(RE +LE)}} \\
    \quad Transformer + 60\% Error
      & 0.513  & 0.535  & 0.651  & 0.592  & 0.534  & 0.658  & 0.665  & 0.662  & 0.470  \\
   \quad \quad\quad\quad\quad\ \ \ \ \  + 90\% Error 
      &\cellcolor{gray!10} \textbf{0.533}  &\cellcolor{gray!10} \textbf{0.563}  &\cellcolor{gray!10} \textbf{0.661}  & \cellcolor{gray!10}\textbf{0.605}  & \cellcolor{gray!10}\textbf{0.544}  & \cellcolor{gray!10}\textbf{0.668}  & \cellcolor{gray!10}\textbf{0.657}  & \cellcolor{gray!10}\textbf{0.674}  & \cellcolor{gray!10}\textbf{0.490}  \\
      \quad 
      &  \cellcolor{gray!10}({\color{ForestGreen}{+20.3\%}}) & \cellcolor{gray!10} ({\color{ForestGreen}{+30.0\%}}) &  \cellcolor{gray!10}({\color{ForestGreen}{+13.2\%}}) &  \cellcolor{gray!10}({\color{ForestGreen}{+15.9\%}}) &  \cellcolor{gray!10}({\color{ForestGreen}{+18.3\%}}) & \cellcolor{gray!10} ({\color{ForestGreen}{+10.2\%}}) &  \cellcolor{gray!10}({\color{ForestGreen}{+0.8\%}}) & \cellcolor{gray!10}({\color{ForestGreen}{+16.2\%}}) & \cellcolor{gray!10}({\color{ForestGreen}{+0.9\%}}) \\
     \quad \quad\quad\quad\quad\ \ \ \ \  + 90\% Error 
      & 0.515  & 0.537 & 0.648 & 0.595  & 0.534& 0.660 & 0.657  & 0.660  & 0.474 \\
    \bottomrule
  \end{tabular}
  }
  \vspace{-1.0em}
  \label{tab:1}
\end{table*}

\begin{tcolorbox}[colback=gray!10, colframe=gray!80, title=\textbf{Case Study: Slef-Reflection and Error-Correction During Inference Time}]
\footnotesize
\begin{tabular}{@{}ll@{}}
\textbf{Raw prediction:}        & \texttt{\textcolor{red}{R<reflect>}KYFHNELM+15.995NYVQEC+57.021QFDSETSL\$} \\
\textbf{Post-processed output:} & \texttt{KYFHNELM+15.995NYVQEC+57.021QFDSETSL} \\
\textbf{Ground-truth label:}    & \texttt{KYFHNELM+15.995NYVQEC+57.021QFDSETSL}
\end{tabular}
\end{tcolorbox}

\begin{tcolorbox}[colback=gray!10, colframe=gray!80, title=\textbf{Case Study: Self-Reflection but no-op During Inference time}]
\footnotesize
\begin{tabular}{@{}ll@{}}
\textbf{Raw prediction:}        & \texttt{KDFFTYME\textcolor{red}{<reflect>}E\$} \\
\textbf{Post-processed output:} & \texttt{KDFFTYME} \\
\textbf{Ground-truth label:}    & \texttt{KDFFTYME}
\label{tab:case-study-reflectio2}
\end{tabular}
\end{tcolorbox}

\begin{table*}[t]
  \centering
  \caption{\label{tab:main}
  Comparison among bio-inspired methods and reflection pretrained model in AA precision. 
  }
  \resizebox{1.0\textwidth}{!}{
  \begin{tabular}{l|ccccccccc|c}
    \toprule
    \textbf{Method} & \textit{Mouse} & \textit{Human} & \textit{Yeast} & \textit{M.mazei} & \textit{Honeybee} & \textit{Tomato} & \textit{R.bean} & \textit{Bacillus} & \textit{C.bacteria} & \textbf{Average} \\
    \midrule
    \textbf{Bio. Inspired Methods} \\
    \quad Peaks       & 0.600 & 0.639 & 0.748 & 0.673 & 0.633 & 0.728 & 0.644 & 0.719 & 0.586 & 0.663 \\
    \quad DeepNovo    & 0.623 & 0.610 & 0.750 & 0.694 & 0.630 & 0.731 & 0.679 & 0.742 & 0.602 & 0.673 \\
    \quad PointNovo   & 0.626 & 0.606 & 0.779 & 0.712 & 0.644 & 0.733 & 0.730 & 0.768 & 0.589 & 0.688 \\
    \quad InstaNovo   & 0.703 & 0.636 & 0.691 & 0.712 & 0.660 & 0.732 & 0.711 & 0.739 & 0.619 & 0.689 \\
    \quad Casanovo, Beam = 1   & 0.717 & 0.649 & 0.752 & 0.713 & 0.706 & 0.763 & 0.714 & 0.753 & 0.663 & 0.704 \\
    \quad HelixNovo, Beam = 1  & 0.750 & 0.648 & 0.758 & 0.766 & 0.699 & 0.757 & 0.771 & 0.799 & 0.661 & 0.734 \\
    \textbf{ Reflection Pretrain} \\
    \quad \cellcolor{green!10} 90\% Error, Beam = 1 & \cellcolor{green!10}\textbf{0.792} & \cellcolor{green!10}\textbf{0.752} & \cellcolor{green!10}\textbf{0.809} & \cellcolor{green!10}\textbf{0.790} & \cellcolor{green!10}\textbf{0.744} & \cellcolor{green!10}\textbf{0.822} & \cellcolor{green!10}\textbf{0.817} & \cellcolor{green!10}\textbf{0.826} & \cellcolor{green!10}\textbf{0.737} & \cellcolor{green!10}\textbf{0.788} \\
    \bottomrule
  \end{tabular}
  }
  \vspace{-1.0em}
\end{table*}

\paragraph{Outperforming Bio-Inspired Models Without Domain Modules.}
As shown in Table~\ref{tab:main}, our reflection-pretrained model achieves state of the art in peptide sequencing task, \textbf{surpassing all bio-inspired baselines}, despite using \textbf{no task-specific architectural modules}. Built on a standard Transformer, it leverages the expressive power of reflection-driven Chain-of-Thought reasoning, demonstrating that procedural reasoning can outperform hard-coded biological priors.

\begin{wraptable}{r}{0.42\linewidth}
\vspace{-1.0em}
\centering
\footnotesize 
\setlength{\tabcolsep}{1.8pt} 
\renewcommand{\arraystretch}{0.85} 
\caption{\small Reflection usage during inference. \textbf{Use}: \% sequences with \texttt{<reflect>}; \textbf{Corr.}: \% reflections that corrected errors; \textbf{Same}: \% reflections that retained the original prediction.}
\label{tab:reflection-usage}
\begin{tabular}{@{}lrrr@{}}
\toprule
\textbf{Model} & \textbf{Use } & \textbf{Corr. } & \textbf{Same } \\
\midrule
Baseline  & 0.0\% & -- & -- \\
Reflect FT & 0.0\% & 0.0\%& 0.0\% \\
PT 60\% Error & 2.3\% & 13.6\% & 63.2\% \\
PT 90\% Error & 4.5\% & 12.7\% & 67.8\% \\
\bottomrule
\end{tabular}
\vspace{-1.4em}
\end{wraptable}

\noindent \textbf{Reflection Behavior Case Study.}
As shown in Table of Case studies, the model exhibits \textbf{strong reflection }capabilities during inference. It successfully detects and corrects its own mistakes using \texttt{<reflect>}, and equally important, it chooses to retain correct outputs when no revision is needed. This balance between self-correction and confident continuation highlights the model's ability to exercise self-doubt as well as self-affirmation.

\begin{figure}[h]
    \centering
    \includegraphics[width=0.65\textwidth]{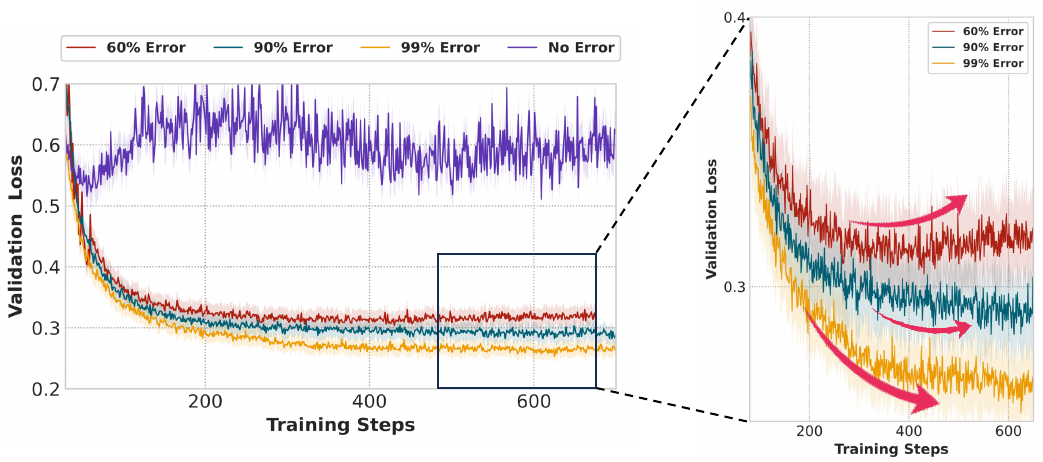} 
    \caption{Error Injection Rate on Validation Loss. Loss Plotting and Logging by Neptune.ai.}
    \label{fig:lossval}
\end{figure}

\noindent \textbf{Impact of Error Injection on Generalization.}
Figure~\ref{fig:lossval} shows the effect of different reflection error ratios on validation loss during pretraining. Without error injection (blue curve), the model exhibits clear signs of overfitting: validation loss initially decreases but begins to rise after 100 steps. In contrast, injecting reflection errors--especially at 90\%  and 99\%  rates--leads to more stable and monotonic decreases in validation loss.
This improvement stems from the fact that our \textbf{batch-dynamic error} alters sequences throughout the training, preventing the model from memorizing fixed input-output mappings. As a result, the model is forced to generalize across perturbed inputs, reducing overfitting.

\paragraph{Human-in-the-Loop Reflection.}
We manually picked 10 sequences where the model failed to detect errors. By manually inserting \texttt{<reflect>} tokens at the error positions and then generating from there, we found that 2 out of 10 sequences were successfully corrected during inference. While this approach is not yet scalable--especially in real-world settings lacking ground-truth labels--it highlights a promising opportunity for expert biologists to intervene and guide reflective reasoning.

Additional results, including evaluation on extended benchmark datasets, the effect of beam size, and further case studies with analysis, are provided in {\color{blue} Appendix}.

\section{Conclusion}
We introduce a reflection-based pretraining approach that equips biological sequence models with intermediate reasoning abilities, inspired by Chain-of-Thought prompting. This improves expressiveness, encourages self-correction, boosts accuracy, and reduces overfitting--particularly for \textit{de novo} peptide sequencing. The framework also enables human-in-the-loop use, bridging biological and natural language models and highlighting the potential of reasoning-driven modeling in biology.


\clearpage
\bibliographystyle{unsrt}
\bibliography{reference}

\clearpage
\appendix
\section{Related Work}

\textbf{Chain-of-Thought Reasoning in Natural Language Models.}
Traditional natural language processing (NLP) systems predominantly followed an \textit{answer-only} generation paradigm, directly mapping inputs to final outputs. This changed with the introduction of Chain-of-Thought (CoT) prompting~\citep{wei2022chain}, which enables models to emit intermediate reasoning steps-non-answer tokens-prior to producing a final answer. CoT significantly improves performance on tasks requiring compositional reasoning~\citep{nye2021show,kojima2022large}, by allowing models to retain transient memory, perform iterative updates, and correct earlier errors during generation. The success of CoT has led to a surge in structured reasoning approaches, including tree-structured~\citep{yao2024tree,long2023large}, graph-based~\citep{besta2024graph,sel2023algorithm}, and decompositional methods~\citep{zhou2022least,drozdov2022compositional,khot2022decomposed}. Further extensions~\citep{zelikman2022star,madaan2024self,suzgun2024meta,ren2025otsurv,ren2025scale,zhang2025supervise,liu2025together,li2025visreason,wei2025ai,pan2025beyond,zhang2025postergen,zhao2025timeseriesscientist,sun2025docagent,liang2025slidegen,xiong2025quantagent} demonstrate that the effectiveness of these methods hinges on the high expressiveness of human language-capable of articulating logic, uncertainty, abstraction, and reflection of which enrich the non-answer space.

\textbf{Limitations of Current Biological Sequence Generation.}
Deep learning models have advanced sequence modeling in biology across DNA~\citep{alipanahi2015predicting,zhang2021deep}, RNA~\citep{yang2022scbert,deng2022rapid}, and proteins~\citep{rives2021biological,jumper2021highly,elnaggar2021prottrans,lin2023evolutionary}. Yet, despite these advances, generation in biological domains-particularly in \textit{de novo} peptide sequencing~\citep{yilmaz2022novo,eloff2023novo,xia2024adanovo}-remains constrained to answer-only prediction. These models operate over token spaces (e.g., 20 amino acids) with inherently limited semantic richness, producing final sequence tokens without externalized reasoning traces. We argue that this limitation is not merely architectural but stems from the low expressive capacity of biological sequence languages. Unlike natural language, biological alphabets are poorly suited for encoding intermediate states, error attribution, or hypothetical exploration capabilities central to CoT-style reasoning.

\textbf{Expressiveness, Computational Power, and Generative Capacity.}
While Transformers with CoT prompting can theoretically simulate Turing-complete machines~\citep{perez2021attention,merrill2023expresssive,strobl2024formal,sun2025ouroboros,you2025uncovering}, their practical reasoning capacity is constrained by input length, memory bandwidth~\citep{arora2009computational,garrison2024memory,you2024calibrating,wen2025route,wen2025beyond}, and crucially, the expressiveness of the output token space. In biological models, the inability to produce diverse non-answer tokens impedes the model's ability to represent internal state transitions or conduct hypothesis-driven exploration. This constitutes a fundamental bottleneck, distinct from issues of model size or architecture. In contrast, natural language models such as GPT~\citep{achiam2023gpt} benefit from an output space rich in semantics, enabling a flexible reasoning substrate. Biological models like ESM~\citep{rives2021biological} or ProLLaMA~\citep{lv2025prollama} lack this capacity, as their generation is grounded in symbolic alphabets that are not optimized for reasoning abstraction.

\textbf{Embedding Reflection and Self-Correction into Biological Models.}
Recent advances in NLP promote self-reflection and critique during inference~\citep{madaan2024self}, but these methods are predominantly post hoc and not integrated into model training. Our work is the first to incorporate reflection and error correction directly into the pretraining phase of biological sequence models. Specifically, we propose a reflection-based pretraining strategy for \textit{de novo} peptide sequencing, where models are exposed to corrupted prediction traces and explicitly trained to identify and amend errors. This goes beyond inference-time prompting, embedding reasoning patterns into model weights. Complementary efforts in scaling Transformer context~\citep{beltagy2020longformer,zhang2023h2o,jiang2023llmlingua} address sequence length limitations but do not address the deeper issue of representational constraints. By introducing a structured set of non-answer tokens and designing tasks that invoke reasoning within biological domains, our framework redefines the generative capacity of biological models to support more intelligent, interpretable, and compositional generation.

\section{Model Expressive Power: From MLPs to Recurrent Architectures and Transformers}

The \textbf{expressive power} of a neural network architecture dictates its fundamental capacity to represent or approximate complex functions, thereby determining its suitability for a diverse range of computational tasks. This capacity is intrinsically linked to the architectural design, especially its mechanisms for information propagation and transformation across processing stages. Architectures with constrained expressive power may falter on tasks demanding deep sequential reasoning or intricate dependency modeling, whereas more expressive models can capture a richer set of complex patterns and computational processes.

\subsection{Defining Effective Computational Depth}

We propose to formalize a critical dimension of expressive power via the concept of \textit{Effective Computational Depth}, denoted $D_{\text{eff}}(\mathcal{M})$. This metric quantifies the maximal length $L$ of a sequence of non-trivial transformations $(\mathcal{T}_i)_{i=1}^L$ that a model $\mathcal{M}$ (with parameters $\theta$) can compose to map an input $x$ to an output $y^*(x)$, potentially averaged over a distribution of tasks $\mathcal{P}_{\text{task}}$. Each $\mathcal{T}_i$ represents a distinct computational step, possibly conditioned on intermediate results generated by preceding transformations $\mathcal{T}_{j<i}$. Formally:

\begin{equation} \label{eq:Deff_smaller}
D_{\text{eff}}(\mathcal{M}) := \sup \left\{ L \in \mathbb{N} \;\middle|\; \exists \mathbf{T}^{(L)} \subseteq \mathcal{M} \text{ s.t. } \left( \forall P \in \mathcal{P}_{\text{task}}^{(L)}, \; \mathcal{R}_{\text{min}}(P, \mathbf{T}^{(L)}) \le \epsilon \right) \right\}
\end{equation}
Here, $\mathcal{P}_{\text{task}}^{(L)}$ signifies tasks inherently requiring $L$ sequential computational steps. $\mathcal{R}_{\text{min}}$ represents the minimum achievable risk (or loss) for task $P$ using the sequence of $L$ transformations $\mathbf{T}^{(L)}$ afforded by model $\mathcal{M}$, and $\epsilon$ is a small tolerance indicating successful task resolution. A higher $D_{\text{eff}}(\mathcal{M})$ signifies a greater capacity for deep sequential processing, a cornerstone of tackling complex computations. The depth complexity, as discussed in prior work, measures the number of sequential steps after considering all parallel processing a model performs. $D_{\text{eff}}(\mathcal{M})$ aims to capture this notion of maximal sequential transformative capacity.

\subsection{Multi-Layer Perceptrons (MLPs)}

Multi-Layer Perceptrons (MLPs) are structured with a fixed number of layers, say $m$. Computation proceeds sequentially through these layers: $h^{(i)} = \sigma(W^{(i)}h^{(i-1)})$. Each layer's operation constitutes a transformation $\mathcal{T}_i$. Since $m$ is a constant, independent of the input sequence length $n$, the effective computational depth $D_{\text{eff}}(\text{MLP})$ is $O(m)$. This effectively becomes $O(1)$ with respect to $n$, as $m$ does not scale with input size. This inherently fixed and often shallow depth restricts MLPs from adeptly solving tasks that necessitate iterative refinement or the processing of sequences with variable and potentially long-range dependencies.

\subsection{Recurrent Neural Networks (RNNs)}

Recurrent Neural Networks (RNNs) incorporate recurrent connections, enabling the output from a previous time step $h_{t-1}$ to be an input to the current step: $h_t = g_{\theta}(h_{t-1}, x_t)$. This architectural feature allows information to persist and be transformed across a number of time steps that can scale with the input sequence length $n$. In this context, each transformation $\mathcal{T}_i$ corresponds to the recurrent computation at a given time step. Consequently, the effective computational depth $D_{\text{eff}}(\text{RNN})$ can reach $O(n)$. This substantial increase in sequential depth capacity empowers RNNs to effectively model sequential data and tasks characterized by temporal dependencies, such as string manipulations or sequential decision-making, far surpassing the capabilities of standard MLPs. RNNs are considered ``recurrence-complete'' as they possess the capability to simulate any one-term recurrent function $h_t = g'(h_{t-1})$, given adequate network capacity and appropriate non-linear activation functions, a property underpinned by the Universal Approximation Theorem.

\subsection{Transformers}

Standard Transformer architectures, despite their considerable success driven by the attention mechanism, operate with a fixed number of layers, $m$, akin to MLPs. While attention allows for parallel processing of tokens within each layer and sophisticated information integration from the entire input sequence $x_{1:t}$, the number of sequential transformation steps (i.e., layers) remains constant. The final layer output $h_t^{(m)}$ is a function of the input $x_{1:t}$ but not directly of a temporally prior hidden state $h_{t-1}^{(m)}$ in the same way as in an RNN.
The transformations $\mathcal{T}_i$ in a Transformer correspond to the operations within each of its $m$ layers. Therefore, the $D_{\text{eff}}(\text{Transformer})$ is $O(m)$, which is $O(1)$ with respect to the input sequence length $n$. This characteristic limits their inherent ability to solve tasks that require a computational depth greater than $m$ without architectural modifications or prompting strategies that simulate deeper recurrence.

\begin{figure}[t]
    \centering
    \includegraphics[width=\linewidth]{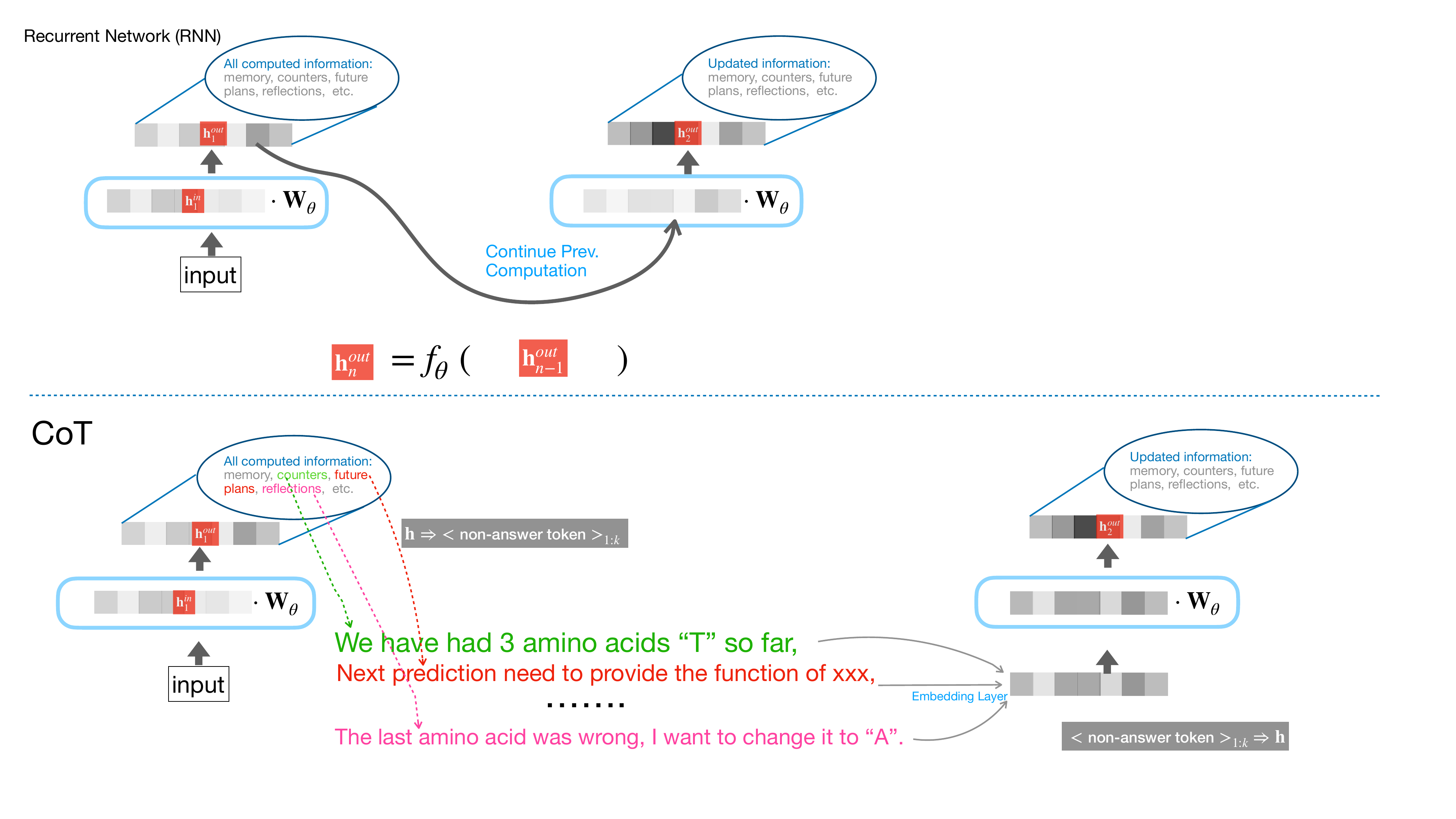}
    \caption{Illustration of how CoT approximate recurrent computation in LMs.}
    \label{fig:cot}
\end{figure}

\section{Chain of Thought: Towards Unbounded Expressive Power in Theory}

The advent of Chain of Thought (CoT) prompting has been empirically shown to significantly enhance the reasoning capabilities of Large Language Models (LLMs). Beyond a mere prompting strategy, CoT can be interpreted from a computational theory perspective as a mechanism that fundamentally reshapes the Transformer's operational dynamics at runtime. In theory, CoT endows autoregressive LLMs with the potential for unbounded expressive power by dynamically altering their computational depth, time complexity, and effective memory.

\subsection{Augmenting Computational Depth via Simulated Recurrence}

Standard Transformer architectures possess a fixed number of layers, $m$, constraining their \textit{Effective Computational Depth}, $D_{\text{eff}}(\text{Transformer})$, to $O(m)$, or $O(1)$ with respect to input sequence length $n$. This architectural limitation inherently restricts their capacity for tasks demanding extensive sequential reasoning steps exceeding $m$.

Chain of Thought circumvents this limitation by simulating a recurrent computational process. The core mechanism involves:
\begin{enumerate}
    \item Encoding the model's internal computational state (hidden state $h$) at the end of a reasoning step into a sequence of natural language tokens $(o_1, o_2, ..., o_k)$ that form the "thought". This can be represented as $h \rightarrow o_{1:k}$.
    \item Appending these tokens to the ongoing input sequence, effectively creating an extended context $x_{n+k} = (x_1, ..., x_n, o_1, ..., o_k)$.
    \item The model then processes this extended sequence, and the CoT string $o_{1:k}$ is decoded back into a computational state $h'$, effectively $o_{1:k} \rightarrow h'$.
\end{enumerate}
This iterative process, $h^{(t)} \rightarrow o^{(t)}_{1:k} \rightarrow h^{(t+1)}$, emulates the recurrent connection $h_t = g(h_{t-1}, x_t)$ found in RNNs, where the model can resume and refine its computation based on the explicitly articulated intermediate state. This transformation allows the internal hidden state $h$, containing comprehensive computational information, to be first externalized into a sequence of "non-answer tokens" (the CoT string). These tokens, which can articulate ongoing reasoning, memory states, or even self-corrections, are then embedded and re-processed by the model to inform subsequent computations, thereby achieving the $o_{1:k} \rightarrow h'$ step. Consequently, the effective computational depth is no longer rigidly tied to the number of physical layers but rather to the length of the generated thought sequence, $T(n)$. The $D_{\text{eff}}(\text{LLM+CoT})$ can thus become $O(T(n))$, theoretically allowing for arbitrarily deep sequential computation, limited primarily by factors like context window length or coherence of the generated thought, rather than by fixed architectural depth. This approximated recurrence is what notably improves the model's computational capacity.

\subsection{Impact on Computational Time}

The simulation of recurrence through CoT inherently impacts computational time. Standard autoregressive generation processes one token at a time. By extending the input sequence with intermediate thought tokens $o_{1:k}$, the CoT process necessitates additional computational steps for both generating these thought tokens and subsequently processing them as part of the augmented input.
If an input of length $n$ requires $T(n)$ CoT steps (where each step might involve generating multiple tokens), the overall time complexity for processing is increased. The time complexity is enhanced to approximately $O(n + T(n)_{\text{tokens}})$, where $T(n)_{\text{tokens}}$ represents the total number of tokens generated as part of the chain of thought. While this increases the computational cost, it is this very extension that facilitates the enhanced reasoning depth.

\subsection{Expanding Effective Memory through Articulated States}

Transformers, through their attention mechanism, can access all previous tokens in their context window. However, the critical information from intermediate computational steps might be diffusely encoded within hidden states and not directly accessible or interpretable for subsequent, distinct reasoning phases that exceed the fixed-layer depth.

CoT provides a mechanism to externalize these intermediate computational states. By articulating the hidden state $h$ (or critical aspects of it, such as memory, counters, or reflections) into natural language strings $o_{1:k}$, these strings act as an explicit, persistent memory trace within the context window. The model can then attend to these textual representations of its prior states in subsequent steps. This allows for a form of "read-write" memory capability where the model "writes" its state as text and "reads" it back to continue computation. This capability is crucial for tasks that require maintaining and manipulating information over extended reasoning chains, effectively emulating more powerful memory structures. For instance, LLMs with CoT can achieve tape-like memory access via attention on the generated text, enabling them to tackle tasks at higher levels of the Chomsky hierarchy, such as context-sensitive tasks, which are typically beyond the reach of standard Transformers without CoT. The expressiveness of natural language is posited to be sufficiently universal to encode diverse types of information, including reasoning states, memory, and intermediate results, which is a critical assumption for this mechanism's success.

In essence, CoT theoretically empowers Transformers by allowing them to dynamically construct deeper computational graphs and manage more explicit memory traces at runtime, thereby transcending some of their inherent architectural limitations.

\section{Why and What's \textit{De Novo} Peptide Sequencing?}

Peptide sequencing from tandem mass spectrometry (MS/MS) data is a cornerstone of proteomics~\citep{aebersold2003mass}, driving advancements from fundamental biological discovery to drug development~\citep{aebersold2003mass,ng2023algorithms}. The goal is to determine the amino acid sequence of peptides directly from their mass spectra. Traditional database search algorithms~\citep{Eng1994,Perkins1999a,Cox2008,Zhang2012}, though widely used, cannot identify novel peptides absent from reference databases. This limitation impedes progress in emerging applications such as \textit{de novo} antibody characterization~\citep{Beslic2022}, neoantigen discovery~\citep{Karunratanakul2019}, and metaproteomics~\citep{Hettich2013}. In contrast, \textit{de novo} peptide sequencing infers sequences without relying on any database, making it indispensable for comprehensive and unbiased proteomic analysis.

The advent of deep learning~\citep{lecun2015deep} has transformed \textit{de novo} sequencing, surpassing traditional algorithmic approaches~\citep{danvcik1999novo,ma2003peaks,frank2005pepnovo}. Early models such as DeepNovo~\citep{tran2017novo} employed CNNs and LSTMs, while more recent efforts have shifted to Transformer-based architectures~\citep{vaswani2017attention}, which offer improved scalability and performance. These models generally fall into two categories: autoregressive (AT) and non-autoregressive (NAT). AT model--such as Casanovo~\citep{yilmaz2022novo,yilmaz2023sequence} and its derivatives (e.g., AdaNovo~\citep{xia2024adanovo}, HelixNovo~\citep{yang2024introducing}, InstaNovo~\citep{eloff2023novo}, ContraNovo~\citep{jin2024contranovo})--predict one amino acid at a time, often achieving high precision through architectural refinements. NAT models like PrimeNovo~\citep{zhang2024pi}, on the other hand, generate sequences in parallel, enabling bidirectional context and faster inference~\citep{gu2017non,xiao2023survey}. Despite this progress, both paradigms largely lack the capacity for explicit intermediate reasoning, which may limit generalization in noisy or uncertain scenarios.

In this work, we propose \ReflectPretrain{}, a novel pretraining strategy that enables models to generate intermediate reasoning steps, bringing CoT-style capabilities to domains beyond natural language. Rather than predicting the final output directly, \ReflectPretrain{} encourages the model to emit structured intermediate tokens that guide and justify the final prediction. By applying this method to the \textit{de novo} sequencing setting, we demonstrate that structured reasoning improves generalization, robustness, and interpretability across peptide prediction tasks.

We identify \textbf{\textit{de novo} peptide sequencing} as an ideal task for exploring structured neural reasoning for five key reasons:
\begin{itemize}
    \item It is a foundational problem in proteomics, as peptide sequencing remains the primary method for identifying naturally occurring amino acid sequences.
    \item The task is inherently reasoning-intensive, requiring interpretation of complex spectral patterns based on biochemical rules, well aligned with the Chain-of-Thought (CoT) paradigm.
    \item Large-scale, high-quality training datasets of spectra paired with peptide sequences are readily available.
    \item Evaluation is exact and unambiguous, with each spectrum matched to a ground-truth sequence, unlike more subjective tasks such as language modeling or structure prediction.
    \item Each predicted token can represent a distinct reasoning step in the sequence inference process, enabling fine-grained supervision and reflection.
\end{itemize}
These properties make the task uniquely suited to evaluate models that go beyond final-answer generation and instead perform stepwise reasoning under uncertainty.

In the \textit{de novo} peptide sequencing task, the model is given a spectrum instance $\mathbf{H} = \{\mathbf{I}, c, m\}$, produced by a mass spectrometer. The spectrum consists of:

\begin{itemize}
    \item $\mathbf{I} = \{(\text{m/z}_1, i_1), (\text{m/z}_2, \text{i}_2), \dots, (\text{m/z}_k, \text{i}_k)\}$, a set of $k$ observed mass-to-charge and intensity pairs, filtered by a signal threshold.
    \item $c \in \mathbb{Z}^+$, the charge state of the precursor ion, and
    \item $m \in \mathbb{R}^+$, the total measured mass of the peptide.
\end{itemize}

The objective is to \textbf{predict the underlying amino acid (protein) sequence} $\mathbf{A} = \{a_1, a_2, \dots, a_n\}$, where each token $a_i \in \mathcal{V}_\textnormal{protein} = \{ \textnormal{20 amino acids tokens}\}$ belongs to the vocabulary of the protein language $\bm{L}_\textnormal{protein} = (\mathcal{G}_\textnormal{protein}, \mathcal{V}_\textnormal{protein})$. The mapping $\mathbf{H} \mapsto \mathbf{A}$ requires the model to reason over the structure and intensity patterns in $\mathbf{I}$ while conforming to biochemical constraints imposed by $c$ and $m$.

\textit{Note that, although we describe our method in the context of the protein sequence space, it applies to other biological sequence prediction tasks-including RNA, DNA, and synthetic polymers}.

\begin{figure}[t]
    \centering
    \includegraphics[width=\linewidth]{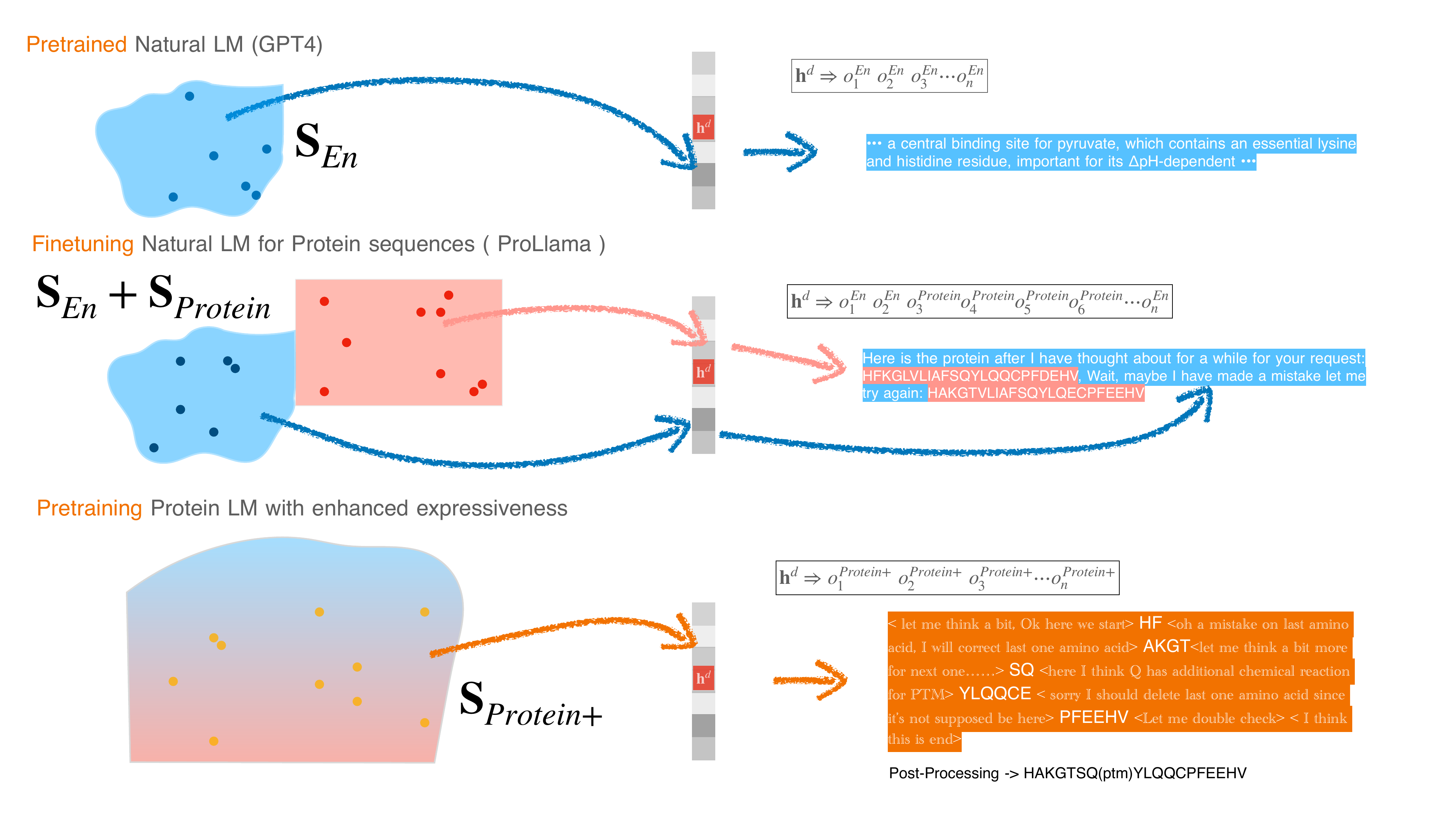}
    \caption{Illustration of the disjoint expressive spaces between natural language (NL) and protein sequences under fine-tuning. CoT reasoning capabilities learned in $\bm{L}_\textnormal{NL}$ do not transfer into $\bm{L}_\textnormal{protein}$, resulting in no shared reasoning ability.}
    \label{fig:disjoint-spaces}
\end{figure}

\section{Why Finetuning Fails to Transfer Reasoning to Protein Sequences}

Let $\bm{L}_\textnormal{NL}$ denote the natural language space (e.g., English), and $\bm{L}_\textnormal{protein}$ denote the protein language space, comprising amino acid tokens with limited compositional structure. Large language models (LLMs) such as LLaMA or GPT-4 are pretrained to reason over $\bm{L}_\textnormal{NL}$ and can perform advanced tasks through CoT prompting--generating intermediate reasoning steps, storing temporary variables, performing error correction, and maintaining self-reflection, all in natural language.

In approaches like \textbf{ProLlama}, a pretrained natural language model is fine-tuned on protein sequences, introducing a new vocabulary $\mathcal{V}_\textnormal{protein}$ and dataset distribution $\mathbb{S}_\textnormal{protein} \subset \mathcal{X}_\textnormal{protein}^*$. While the transformer architecture remains shared, the representations fine-tuned on proteins diverge significantly from those learned in natural language. As illustrated in Figure~\ref{fig:disjoint-spaces}, the model's learned expressiveness splits into two largely non-overlapping subspaces:
\begin{equation}
\mathbb{S}_\textnormal{NL} \cap \mathbb{S}_\textnormal{protein} \approx \emptyset
\end{equation}
This results in a model that performs well on surface-level protein generation but cannot leverage its natural language reasoning skills--because those abilities remain confined to the $\bm{L}_\textnormal{NL}$ space and are inaccessible in $\bm{L}_\textnormal{protein}$.

To understand this intuitively, consider a bilingual LLM pretrained on English. If we fine-tune it on Chinese texts without joint training or alignment, the model learns to generate Chinese tokens fluently, but cannot reason about Chinese using English thinking patterns. For example, tokens like "let me double check" or "I made a mistake" have no meaningful grounding in the Chinese token space unless the model was trained to use them cross-lingually. The model ends up with two islands of knowledge--English and Chinese--that cannot communicate or share reasoning capacity.

The same phenomenon occurs in protein modeling. ProLlama retains its CoT reasoning ability in English but cannot express or apply that reasoning in $\bm{L}_\textnormal{protein}$. As a result, it cannot generate intermediate reasoning tokens like \texttt{<reflect>} or revise outputs mid-generation--capabilities that are essential for self-correction, debugging, or guided sequence exploration.

In contrast, our reflection-pretraining approach explicitly introduces auxiliary reasoning tokens $\mathcal{V}_\textnormal{reflect}$ and trains the model to use them within the protein generation process. This expands the language from $\bm{L}_\textnormal{protein}$ to an augmented form:
\begin{equation}
\bm{L}_\textnormal{protein+} = (\mathcal{G}_\textnormal{protein}, \mathcal{V}_\textnormal{protein} \cup \mathcal{V}_\textnormal{reflect})
\end{equation}
and correspondingly enlarges the expressive space from $\mathbb{S}_\textnormal{protein}$ to $\mathbb{S}_\textnormal{protein+}$, allowing the model to generate and interpret internal reasoning sequences natively in the protein domain:
\begin{equation}
|\mathbb{S}_\textnormal{protein+}| \gg |\mathbb{S}_\textnormal{protein}|
\end{equation}

\section{Experiments}
\textbf{Model Training.}
We adopt the same architecture and optimization configuration as in prior work of training Transformer for De novo Peptide Sequencing. All inputs--including MS/MS peaks, precursor features, and amino acid tokens--are first embedded into a shared 400-dimensional latent space.

The model is built on a 9-layer Transformer, with each layer comprising 8 attention heads and a 1024-dimensional feedforward network. This architecture forms the backbone for both the encoder and decoder components.

Training is conducted with a batch size of 1600 and an initial learning rate of 4e-4. The learning rate linearly warms up during the first epoch before decaying via a cosine schedule. Optimization is performed using AdamW~\cite{kingma2014adam}, and all models are trained for 30 epochs on eight A100 GPUs.

These hyperparameters--embedding dimension, transformer depth, attention head count, and learning rate policy--are kept fixed across all experiments unless explicitly stated. Additional implementation details can be found in our released codebase.

\textbf{Base Bio-inspired Model Designs.} We evaluate our approach against a range of bio-inspired peptide sequencing models, each of which incorporates domain-specific knowledge or architectural modifications tailored to biological data:

\begin{itemize}
    \item \textbf{Peaks}~\cite{ma2003peaks} employs a protein database search strategy that combines tandem mass spectrometry with homology-based alignment. This approach excels in navigating ambiguity in peptide identification using prior biological information.
    
    \item \textbf{DeepNovo}~\cite{tran2017novo} combines convolutional and recurrent neural networks with peptide-specific preprocessing and adopts a species-wise cross-validation scheme.
    
    \item \textbf{PointNovo}~\cite{qiao2021computationally} builds on point-cloud processing ideas to handle varied-resolution spectra without increasing model complexity, showcasing robustness through biologically-informed design.

    \item \textbf{InstaNovo}~\cite{eloff2023novo} introduces a diffusion model that improves performance through iterative refinement of predicted sequences.
    
    \item \textbf{HelixNovo}~\cite{yang2024introducing} extends CasaNovo by modifying the MS/MS input to explicitly encode complementary b/y ion relationships--information that is biologically grounded in peptide fragmentation rules.
    
\end{itemize}

In contrast to these models, our reflection-pretrained Transformer introduces no biological heuristics or domain-specific preprocessing. Instead, it augments a standard model's reasoning capacity via token-level reflection and self-correction. This purely cognitive enhancement allows the model to outperform bio-inspired baselines without incorporating any explicit biochemical priors.

\begin{tcolorbox}[colback=gray!10, colframe=gray!80, title=\textbf{Case Study: Manual Reflection Improves Prediction}]
\label{human}
\footnotesize
\begin{tabular}{@{}ll@{}}
\textbf{Raw prediction:}        & \texttt{RLANLYWL\$} \\
\textbf{Manual intervention:}   & \texttt{RL\textcolor{red}{<reflect>}} \\
\textbf{Output after intervention:} & \texttt{RL\textcolor{red}{<reflect>}MNFYGFL} \\
\textbf{Ground-truth label:}    & \texttt{RMNFYGFL}
\end{tabular}
\end{tcolorbox}

\paragraph{Human-Guided Reflection for Correction.}
Case study in Table \ref{human} demonstrates the potential of human-in-the-loop intervention using reflection tokens. As shown above, the model initially produces an incorrect prediction (``RLANLYWL''), but upon manually inserting a \texttt{<reflect>} token after the first residue, the model self-corrects and generates the correct target sequence (``RMNFYGFL''). 

Notably, the correction cascades beyond a single token--the model adjusts not just the second residue but the entire downstream sequence. This highlights the model's capacity for global reasoning triggered by a local reflective signal. Although we do not yet have automated methods for locating such errors in unlabeled real-world data, this opens the door for expert-guided inference: domain experts can flag uncertain regions and inject \texttt{<reflect>} tokens to improve model output. This enables a promising direction for collaborative, controllable biological sequence generation.

\paragraph{Analysis of Beam Size Influence.}
\label{sec:beam_size_analysis}

We examined the effect of beam search width on the performance of the Reflection Pretrain model. As shown in Table~\ref{tab:2}, increasing the beam size consistently improves model performance across both Amino Acid (AA) Precision and Peptide Recall metrics. Specifically, AA Precision increased from $0.788$ at beam size 1, to $0.797$ at beam size 3, and further to $0.806$ at beam size 5. This reflects an absolute improvement of $0.018$, corresponding to a relative increase of approximately $2.28\%$. A similar trend was observed for Peptide Recall, which rose from $0.600$ (beam size 1) to $0.608$ (beam size 3), and reached $0.617$ at beam size 5--an absolute gain of $0.017$ and a relative improvement of approximately $2.83\%$. These results suggest that larger beam sizes enable the model to explore a broader hypothesis space during decoding, thereby reducing the risk of prematurely discarding promising sequence candidates. Nevertheless, the performance benefits must be balanced against the increased computational cost in both time and memory--an inherent trade-off in beam search.

\begin{table*}[t]
  \centering
  \caption{
  Comparison among bio-inspired methods and reflection reflection-pretrained model. 
  }
  \label{tab:2}
  \resizebox{1.0\textwidth}{!}{
  \begin{tabular}{l|ccccccccc|c}
    \toprule
    \textbf{Method} & \textit{Mouse} & \textit{Human} & \textit{Yeast} & \textit{M.mazei} & \textit{Honeybee} & \textit{Tomato} & \textit{R.bean} & \textit{Bacillus} & \textit{C.bacteria} & \textbf{Average} \\
    \midrule
    \multicolumn{11}{c}{\textbf{AA Precision}} \\
    \textbf{Bio. Inspired Methods} \\
    \quad Peaks       & 0.600 & 0.639 & 0.748 & 0.673 & 0.633 & 0.728 & 0.644 & 0.719 & 0.586 & 0.663 \\
    \quad DeepNovo    & 0.623 & 0.610 & 0.750 & 0.694 & 0.630 & 0.731 & 0.679 & 0.742 & 0.602 & 0.673 \\
    \quad PointNovo   & 0.626 & 0.606 & 0.779 & 0.712 & 0.644 & 0.733 & 0.730 & 0.768 & 0.589 & 0.688 \\
    \quad InstaNovo   & 0.703 & 0.636 & 0.691 & 0.712 & 0.660 & 0.732 & 0.711 & 0.739 & 0.619 & 0.689 \\
    \quad Casanovo, Beam = 1   & 0.717 & 0.649 & 0.752 & 0.713 & 0.706 & 0.763 & 0.714 & 0.753 & 0.663 & 0.704 \\
    \quad HelixNovo, Beam = 1  & 0.750 & 0.648 & 0.758 & 0.766 & 0.699 & 0.757 & 0.771 & 0.799 & 0.661 & 0.734 \\
    \textbf{ Reflection Pretrain} \\
    \quad \cellcolor{green!10} 90\% Error, Beam = 1 & \cellcolor{green!10}\textbf{0.792} & \cellcolor{green!10}\textbf{0.752} & \cellcolor{green!10}\textbf{0.809} & \cellcolor{green!10}\textbf{0.790} & \cellcolor{green!10}\textbf{0.744} & \cellcolor{green!10}\textbf{0.822} & \cellcolor{green!10}\textbf{0.817} & \cellcolor{green!10}\textbf{0.826} & \cellcolor{green!10}\textbf{0.737} & \cellcolor{green!10}\textbf{0.788} \\
    \quad \cellcolor{green!10} 90\% Error, Beam = 3 & \cellcolor{green!10}\textbf{0.799} & \cellcolor{green!10}\textbf{0.763} & \cellcolor{green!10}\textbf{0.818} & \cellcolor{green!10}\textbf{0.800} & \cellcolor{green!10}\textbf{0.755} & \cellcolor{green!10}\textbf{0.827} & \cellcolor{green!10}\textbf{0.832} & \cellcolor{green!10}\textbf{0.836} & \cellcolor{green!10}\textbf{0.745} & \cellcolor{green!10}\textbf{0.797} \\
    \quad \cellcolor{green!10} 90\% Error, Beam = 5 & \cellcolor{green!10}\textbf{0.805} & \cellcolor{green!10}\textbf{0.774} & \cellcolor{green!10}\textbf{0.826} & \cellcolor{green!10}\textbf{0.810} & \cellcolor{green!10}\textbf{0.766} & \cellcolor{green!10}\textbf{0.831} & \cellcolor{green!10}\textbf{0.846} & \cellcolor{green!10}\textbf{0.846} & \cellcolor{green!10}\textbf{0.753} & \cellcolor{green!10}\textbf{0.806} \\
    \midrule
    \multicolumn{11}{c}{\textbf{Peptide Recall}} \\
    \textbf{Bio. Inspired Methods} \\
    \quad Peaks       & 0.197 & 0.277 & 0.428 & 0.356 & 0.287 & 0.403 & 0.362 & 0.387 & 0.203 & 0.333 \\
    \quad DeepNovo    & 0.286 & 0.293 & 0.462 & 0.422 & 0.330 & 0.454 & 0.436 & 0.449 & 0.253 & 0.387 \\
    \quad PointNovo   & 0.355 & 0.351 & 0.534 & 0.478 & 0.396 & 0.513 & 0.511 & 0.518 & 0.298 & 0.439 \\
    \quad InstaNovo   & 0.471 & 0.455 & 0.559 & 0.528 & 0.466 & 0.732 & 0.564 & 0.576 & 0.416 & 0.530 \\
    \quad Casanovo    & 0.443 & 0.433 & 0.584 & 0.522 & 0.460 & 0.606 & 0.652 & 0.580 & 0.413 & 0.521 \\
    \textbf{Reflection Pretrain} \\
    \quad \cellcolor{green!10}90\% Error, Beam=1 & \cellcolor{green!10}\textbf{0.533} & \cellcolor{green!10}\textbf{0.563} & \cellcolor{green!10}\textbf{0.661} & \cellcolor{green!10}\textbf{0.605} & \cellcolor{green!10}\textbf{0.544} & \cellcolor{green!10}\textbf{0.668} & \cellcolor{green!10}\textbf{0.657} & \cellcolor{green!10}\textbf{0.674} & \cellcolor{green!10}\textbf{0.490} & \cellcolor{green!10}\textbf{0.600} \\
    \quad \cellcolor{green!10}90\% Error, Beam=3 & \cellcolor{green!10}\textbf{0.540} & \cellcolor{green!10}\textbf{0.573} & \cellcolor{green!10}\textbf{0.669} & \cellcolor{green!10}\textbf{0.615} & \cellcolor{green!10}\textbf{0.555} & \cellcolor{green!10}\textbf{0.674} & \cellcolor{green!10}\textbf{0.667} & \cellcolor{green!10}\textbf{0.685} & \cellcolor{green!10}\textbf{0.497} & \cellcolor{green!10}\textbf{0.608} \\
    \quad \cellcolor{green!10}90\% Error, Beam=5 & 
    \cellcolor{green!10}\textbf{0.546} & \cellcolor{green!10}\textbf{0.582} & \cellcolor{green!10}\textbf{0.676} & \cellcolor{green!10}\textbf{0.624} & \cellcolor{green!10}\textbf{0.565} & \cellcolor{green!10}\textbf{0.680} & \cellcolor{green!10}\textbf{0.676} & \cellcolor{green!10}\textbf{0.696} & \cellcolor{green!10}\textbf{0.504} & \cellcolor{green!10}\textbf{0.617} \\
    \bottomrule
  \end{tabular}
  }
  \vspace{-1.0em}
\end{table*}

\section{Limitations}
Although our proposed reflection-based pretraining paradigm marks a significant step forward in equipping biological sequence models with reasoning capabilities, it is not without limitations. First, the approach depends heavily on the availability of high-quality reflection datasets, which are both challenging and time-consuming to curate for specific biological tasks. Second, the computational cost of pretraining with reflection mechanisms is higher than that of traditional methods, as the inclusion of non-answer tokens increases sequence length and training complexity. Lastly, while our framework facilitates human-in-the-loop interactions, its practical adoption in real-world biological workflows hinges on biologists' ability to interpret and effectively utilize the intermediate reasoning outputs, which may require additional training or domain expertise. These limitations highlight several avenues for future research, including improved dataset curation, optimization of computational efficiency, and the development of more user-friendly interfaces to enhance the accessibility and impact of biological reasoning models.

\section{Broader Impact}
The capacity to imbue biological sequence models with reasoning and self-correction, as demonstrated for de novo peptide sequencing, holds transformative potential for society, primarily by accelerating biomedical discovery for new drugs and therapies, and advancing biotechnology through the design of novel enzymes and materials. This advancement is not without significant ethical responsibilities, chief among them mitigating dual-use risks related to the creation of harmful biological agents, ensuring the safety of AI-designed sequences, and promoting equitable access to these powerful tools. Consequently, the path forward requires a concerted effort towards responsible innovation, incorporating robust safety protocols, ethical guidelines, and mechanisms for human oversight-such as the human-in-the-loop generation. Our work supports harnessing these capabilities for broad societal benefit while safeguarding against potential harms.

\end{document}